# Text Data Augmentation Made Simple
# By Leveraging NLP Cloud APIs

**Claude COULOMBE**
Doctorant Informatique Cognitive
TÉLUQ / UQAM
coulombe.claude@univ.teluq.ca
Consultant
Lingua Technologies Inc. / DataFranca

## ABSTRACT

In practice, it is common to find oneself with far too little text data to train a deep neural network. This "Big Data Wall" represents a challenge for minority language communities on the Internet, organizations, laboratories and companies that compete the GAFAM (Google, Amazon, Facebook, Apple, Microsoft).

While most of the research effort in text data augmentation aims on the long-term goal of finding end-to-end learning solutions, which is equivalent to "using neural networks to feed neural networks", this engineering work focuses on the use of practical, robust, scalable and easy-to-implement data augmentation pre-processing techniques similar to those that are successful in computer vision.

Several text augmentation techniques have been experimented. Some existing ones have been tested for comparison purposes such as noise injection or the use of regular expressions. Others are modified or improved techniques like lexical replacement. Finally more innovative ones, such as the generation of paraphrases using back-translation or by the transformation of syntactic trees, are based on robust, scalable, and easy-to-use NLP Cloud APIs.

All the text augmentation techniques studied, with an amplification factor of only 5, increased the accuracy of the results in a range of 4.3% to 21.6%, with significant statistical fluctuations, on a standardized task of text polarity prediction. Some standard deep neural network architectures were tested: the multilayer perceptron (MLP), the long short-term memory recurrent network (LSTM) and the bidirectional LSTM (biLSTM). Classical XGBoost algorithm has been tested with up to 2.5% improvements.

# 1. INTRODUCTION

In practice, it is common to find oneself with far too little data to train a deep neural network. This "Big Data Wall"[1] is a well-kept secret, as is the enormous amount of computation required to properly train the models and the delicate development of neural network architectures that requires empirical know-how that is still poorly understood and documented.

This is particularly relevant for natural language processing (NLP) applications, where the "Big Data Wall" represents a challenge for minority language communities on the Internet which have small electronic corpora, organizations, laboratories and companies that compete with the giant web companies, the GAFAM (Google, Amazon, Facebook, Apple, Microsoft).

The need for a large amount of data is not specific to deep learning, it is actually related to the complexity of the task to be solved. Unsurprisingly, learning a complex task requires a complex model and a greater amount of training data.

The main benefit of deep learning over conventional machine learning is that deep learning makes it easier to build complex statistical models, with little or no features engineering, because deep learning is reknown to be able to learn directly features from the data in a hierarchical and distributional way [Goodfellow, Bengio & Courville, 2016].

This work deals with text data augmentation. Data augmentation is a poor term, somewhat imprecise, because it is rather an amplification, since it starts from existing data to create new ones, while preserving the meaning that must remain invariant. The idea of "semantically invariant transformation" is at the heart of the text data augmentation process. The results of data augmentation is also called synthetic data, generated data, simulated data or artificial data.

Data augmentation can also be considered as a regularization technique since that is used to avoid overfitting.

Although the studied techniques are applying to all types of texts, this work focuses on sentence augmentation.

---

[1] Or the « Big Data Barrier »

## 2.    RELATED WORKS

In computer vision, that's common practice to create new images by geometric transformations which preserve the similarity. In 1997, [Ha & Bunke, 1997] created additional images data by applying transformations to real images in order to help recognition of handwriting characters. These are isometries (distance preserving) such as translations, rotations, when appropriate reflections or flipping over an orthogonal axis, and homotheties (scaling) such as enlargements / reductions (rescaling) and even elastic deformations [Simard, Steinkraus & Platt, 2003]. The intensity of RGB (color coding) channels can also be altered by using principal component analysis to generate new data [Krizhevsky, Sutskever & Hinton, 2012]. Such data augmentation has been used extensively to win the ImageNet competition in 2012 with a huge 60 millions parameters neural networks.

In speech recognition, data augmentation is achieved by manipulating the signal as slowing or accelerating it [Ko et al, 2015], noise injection, and spectrogram modification [Jaitly & Hinton, 2013].

Unlike computer vision or speech data, text data has not many popular techniques for data augmentation. Until recently, the only widespread text augmentation technique was the lexical substitution which consists to replace a word by its synonym using a thesaurus [Zhang & LeCun, 2015]. Even recently usage of data augmentation for NLP has been very limited [Kobayashi, 2018].

It is legitimate to ask why? First, because natural language data are difficult to process. According to [Goldberg, 2017], text data are difficult to process because they are symbolic, discrete, compositional and sparse. To this it must be added that text data are hierarchical, noisy, full of exceptions and ambiguous. Second, gradient descent-based learning techniques do not apply directly to discrete data like text as explained by Ian Goodfellow in his comments on Reddit about the difficulty of text GAN [Goodfellow, 2016]. Third, it's harder to generate realistic textual data. For instance, conventional autoencoders (AE) failed to generate realistic sentences since they do not constraint the latent space and work word by word [Bowman et al, 2015]. Fourth, everyone is doing it, yet nobody ever talk about it.

That said, it is an open secret that usage of hand-crafted rules which includes noise injection and regular expressions would commonly be used by practioners to augment text data, a bit like hand-crafted rules found in almost all chatbot engines. A nice practical example is the Python library NoiseMix [Bittlingmayer, 2018] which proceeds by injection of textual noise.

Maybe the explanation is more sociologic than technical. Most of the research effort in text data augmentation focuses on the noble task of finding end-to-end learning solutions or building generative models, which is often equivalent to "using neural networks to feed neural networks".

For instance, various attempts have been made to develop brand new deep generative models: variational autoencoders (VAE) that work on entire sentences [Bowman et al, 2015], VAE with disentangled latent representations and attributes discriminators (tense and sentiment) [Hu et al, 2017], special generative adversarial network (GAN) called TextGAN that combines LSTM (generator) and CNN (discriminator) with soft word-vector space [Zhang et al, 2017], with target syntactic template [Iyyer et al, 2018] in order to control syntactic variations, SeqGAN which combines reinforcement learning (RL) with GAN [Yu et al, 2017], Neural Machine Translation [Edunov & al, 2018], [Prabhumoye et al., 2018], [Wieting, Mallinson & Gimpel, 2017] and even chinese poetry generator [Rajeswar et al, 2017]. Furthermore, new lexical replacement techniques have been developed but essentially based on deep neural networks [Kobayashi, 2018], [Wang et al, 2018], [Fadaee, Bisazza & Monz, 2017]. There is a platform, called TexyGen, to benchmark different text GAN generative models [Zhu et al, 2018], [Texygen, 2018].

All those are fine and essential long-term research works, but in the meantime there is an urgent need for short term practical solutions. This piece of work, which is more an engineering endeavour than a fundamental research, is in the vein of the methods used in computer vision which consist of applying invariant transformations to original data in order to generate augmentation data [Simard, Steinkraus & Platt, 2003]. That is a way of working which goes back to around 2000. It is not even appropriate to compare the two approaches since both pursue different goals at different time scales. Eventually the question also arises as to the computational effectiveness of  deep learning generative model for data augmentation. Obviously, one has to put in the balance the cost of training the models and the cost of using the models once trained. At the end of the day, the results should speak for themselves.

## 3.    METHOD

### 3.1 Basic Assumption

According to [Zhang & LeCun, 2015] the use of paraphrases would be ideal, but the authors did not envisage doing so in an automatic way, except by the technique of replacing words using a thesaurus.

Our goal is to reproduce in NLP the same augmentation techniques used with great success in artificial vision that consist of data transformations mainly at the data pre-processing stage. Our basic scientific assumption is that textual data augmentation techniques can be derived from simple, practical and easy-to-use natural language processing (NLP) and machine learning. which facilitate the training of large statistical models. In addition, we are interested to take advantage of the NLP Cloud APIs available today.

If data is scarce and the original data distribution has transformation invariance properties, generating additional data using transformations can improve performance [Simard, Steinkraus & Platt, 2003], [Yaeger, Lyon , & Webb, 1997].

## 3.2 Attempt to formalize data augmentation

Statistically, the additional data generated should as far as possible be distributed according to the same statistical distribution as the original data.

| Rule of respect the statistical distribution |
| --- |
| *The augmented data must follow a statistical distribution similar to that of the original data.* |

On the semantic level, the idea is to find transformations that will not affect the meaning of the data but that will contribute to the learning of "new forms" in the sense of pattern recognition. This is complicated when one considers the golden rule that a human being must judge "plausible" the amplified data.

| Golden Rule of Plausibility |
| --- |
| *A human being should not be able to distinguish between the amplified data and the original data.*<br>[Géron, 2017b] |

Once a plausible transformation has been chosen, applying it to a given data is simple, but the inverse problem (transformation invariance) of finding the meaning of the data from the transformed data can be very difficult. Precisely machine learning algorithms are very effective for solving inverse problems [Simard, Steinkraus & Platt, 2003].

Data augmentation is easier in supervised classification situations involving simple target tags. A classifier takes as input a large vector X and associates it with a single label of the target class y of small size. This means that the classifier must be invariant to a wide variety of transformations on the input data X but the target label remains simple. It is potentially easier to generate new pairs (X, y) simply by transforming the many X entries in the training data set while retaining the target tag y [Goodfellow, Bengio & Courville, 2016].

The most recognized techniques for natural language processing use supervised machine learning algorithms. From a supervised classification task, transformations can be applied to generate additional data and allow the learning algorithm to induce transformation invariance. This invariance is integrated into the model parameters at the time of learning. It becomes somewhat free when one applies the model (at the time of the inference) to new data since the parameters of the model are fixed [Simard, Steinkraus & Platt, 2003].

Mathematically, a subset $E$ of the domain $U$ of a transformation is an invariant set for this transformation when $x \, \varepsilon \, E \;\; \Rightarrow \;\; T(x) \; \varepsilon \, E$

In an effort of formalization, we can affirm what we will call the rule of semantic invariance. The augmentation of the data is an invariant transformation on the meaning, or an invariant transformation for meaning or more simply a semantically invariant transformation. This is the term we will remember in the end.

| Semantic invariance rule |
| --- |
| *Data augmentation involves semantically invariant transformations.* |

For example, in supervised learning, allowed transformations are those that will not change the class label of the new data generated (label-preserving transformations). For example, to differentiate between the letter "b" and the letter "d" or between the number "6" and the number "9", horizontal reflection and rotation of 180º would not be allowed transformations for character recognition [Goodfellow, Bengio & Courville, 2016].

| Semantic invariance rule in supervised learning |
| --- |
| *In supervised learning, the transformations allowed for data augmentation are those that do not modify the class label of the new data generated.* |

By carefully combining the proposed transformations, one can at the same time obtain paraphrases more and more distant from the original sentence and retain its meaning. However, we must remain vigilant because each transformation runs the risk of moving away more and more from the original meaning. Hence the rule of thumb of the telephone game.

The telephone game (Chinese Whispers) is a children's game where the goal is to circulate a phrase of word of mouth in a low voice through a line of players. The first player in the queue invents a sentence and the last player in the queue recites the sentence that was given to him. The interest of the game is to compare the final version of the sentence to its original version.

| Telephone Game Rule of Thumb |
|---|
| *In order to respect the semantic invariance, the number of successive or combined transformations must be limited, empirically to two (2).* |

### 3.3 Technique 1 - "Textual Noise" Injection

What comes closest to a continuous change in a text is the injection of weak textual sounds: changes, additions, deletions of letters in words, change of case, modification of punctuation.

The injection of noise into a neural network can be considered as a form of data augmentation. It is possible to improve the robustness of a neural network by adding random noise to its inputs. In this sense, dropout, a powerful regularization technique, can be considered as a process of data augmentation by noise [Goodfellow, Bengio & Courville, 2016].

We hesitate to call noise injection a text data augmentation because the addition of noise generally contributes more to the robustness of learning [Xie et al, 2017] than to the recognition of new forms in the data. We can very well discuss it, but since the rule of invariance applies, the low noise injection will be considered as an augmentation and subject to the experiment that will conclude.

| Textual noise injection |
|---|
| *Light textual noise injection is a semantically invariant transformation.* |
| *Strong textual noise injection is not a semantically invariant transformation.* |

This raises the question of the frontier between light noise and chaos. It might be interesting to investigate this topic by controlling the amount and kind of textual noise injected and to observe the results.

## 3.4 Technique 2 - Spelling Errors Injection

The idea is to generate texts containing common misspellings in order to train our models which will thus become more robust to this particular type of textual noise.

The spelling injection algorithm is based on a list of the most common mispellings in English. This list was compiled by the publisher of the Oxford Dictionaries [Oxford Dictionaries, 2018].

| Spelling Errors Injection |
| --- |
| *Spelling errors injection is a semantically invariant transformation.* |

Again, it would be interesting to measure experimentally the effect of the gradual injection of more and more spelling errors.

## 3.5 Technique 3 - Word Replacement using thesaurus

Lexical replacement consists of proposing one or more words that can replace a given word. These words are typically true synonyms of this word.

Generally, there is no replacement for grammatical words. Here, in order of increasing difficulty, the types of words that are candidates for lexical substitution: adverbs, adjectives, nouns and verbs. Verbs replacement is particularly challenging because of the different arguments that accompany the verbs. In many situations, we limit to replace only adverbs and adjectives, sometimes we add nouns, more rarely verbs.

For the lexical replacement, one will favor the use of hyperonyms (more general word, tulip => flower) and one will avoid the use of hyponyms (word more precise, flower => tulip).

| Lexical Replacement Rules of Thumb |
| --- |
| *Replacing a word by a real synonym is a semantically invariant transformation.* |
| *Replacing a word by a hyperonym (more general word)*<br>*is a semantically invariant transformation.* |

> *Replacing a word by a hyponym (more specific word)*
> *is usually not a semantically invariant transformation.*

> *Replacing a word by an antonym is not a semantically invariant transformation.*

A first approach for lexical substitution [Zhang & LeCun, 2015] uses a thesaurus like Wordnet. It is worth remembering that Wordnet-type language resources [Miller & al, 1990] have been painstakingly crafted by hand. We have to deal with the inventory of the particular senses chosen by those who created these resources. Typically, the algorithm generates all possible synonyms candidates then filter them according to various criteria .

The main difficulty of lexical replacement comes from the ambiguity of the natural language. When a word has more than one meaning, it has several different synonyms.

A thesaurus like WordNet is built to associate to each dictionary entry a list that contains several sets of synonyms or synsets. Each synsets} corresponds to a particular meaning. The number of meanings is variable and depends most often on a human annotation that can be more or less exhaustive.

The challenge is to choose the right synset. Several strategies can be used to find the right set of synonyms. For example, one can choose the most common meaning based on the number of occurrences in a reference corpus. One can also use the context of the word and the information which accompanies each synset in the dictionary like definitions, examples, to compute a similarity measure between the context of the word and the companion information of each synset and finally choose the most similar. We have opted for this latter strategy.

Sometimes, it happens that some thesaurus returns antonyms among synonyms. In that case, it can be required to filter the synonyms output using a dictionary of antonyms.

## 3.6 Text augmentation by paraphrases generation

Continuing to explore further text augmentation techniques, we come to conceive as [Zhang & LeCun, 2015] that the augmentation of text data ideally passes through the generation of paraphrases.

By definition, paraphrase is an alternative surface form in the same language which expresses the same semantic content as the original form [Madnani & Dorr, 2010]. Paraphrases may occur at several levels. For instance, words having the same meaning, which are commonly referred to synonyms, can be also considered to as lexical paraphrases. There are paraphrases at the level of the group of words or phrase (phrasal paraphrase), for instance « take over » and « assume control of », also at the level of the complete sentence (sentential paraphrase), like with « I finished my work » and « I completed my assignment. » [Madnani & Dorr, 2010].

| Definition of the perfect paraphrase |
|---|
| *In addition to being meaning-preserving, an ideal paraphrase must also diverge as sharply as possible in form from the original while still sounding natural and fluent.* <br> [Chen & Dolan, 2011] |

Anyone familiar with natural language processing knows that it should be possible to generate paraphrases using grammars.

## 3.7 Technique 4 - Paraphrases generation using regular expressions

The first type of transformation that comes to mind are surface transformations. A surface transformation is a transformation that ignores syntax and is done with simple pattern matching rules. Regular expressions (regex) are powerful tools to manage transformations based on pattern matching.

The surface transformations that can be produced with regular expressions are to be preferred because they are simple and very efficient in terms of computation.

Many of the surface transformations are dependent on the language being processed; as for example, contractions in English. However, there are surface transformations common to several languages such as noise injection, the injection of particular types of spelling errors, transformations on culturally shared objects such as dates, location, entity names, units of measurement.

In the same vein, there are many transformations at the level of abbreviations, acronyms, notations and orthographic variants. There are potentially thousands of rules that can be partly made more general by factoring their behavior.

For example, in English, the transformation of a verbal form into a contracted form (contraction) and its inverse (expansion) is a semantically invariant transformation provided that any ambiguity that can lead to a potential misinterpretation is not resolved. Obviously this happens in the context of mechanical and local transformations where there is no way to resolve ambiguities.

| Examples of a text surface transformation |
|---|
| *The transition to a contracted verbal form and its inverse is a semantically invariant transformation provided that the ambiguities are preserved.* |

```
I am => I'm, you are => you're, he is => he's, it is => it's, she is => she's, we are => we're, they are
=> they're,, I have => I've, you have => you've, we have => we've, they have => they've, he has => he's,
it has => it's, she has => she's, I will => I'll, you will => you'll, he will => he'll, are not =>
aren't, is not => isn't, was not => wasn't, ..., I'm => I am, I'll => I will, you'll => you will, he'll
=> he will, aren't => are not, isn't => is not, wasn't => was not, weren't => were not, couldn't =>
could not, don't => do not, doesn't => does not, didn't => did not, mustn't => must not, shouldn't =>
should not, can't => can not, can't => cannot, won't => will not, shan't => shall not
```

In order to preserve the semantic invariance, it is allowed to introduce ambiguities but it is forbidden to resolve ambiguities that could lead to misinterpretation. For example the transformations "he is" to "he's" and "he has" to "he's" will be allowed even if they introduce ambiguous sentences. But the inverse transformations from "he's" to "he is" and "he's" to "he has" are forbidden because they could introduce a misinterpretation by solving an ambiguity without justification.

| Examples of transformations to avoid because they resolves an ambiguity without justification |
|---|
| *she's => she is*<br>*she's => she has* |

| The « respect for ambiguity » rule of thumb |
|---|
| *A transformation that create ambiguity or imprecision is often considered semantically invariant.* |
| *A transformation that resolves an ambiguity, by specifying an information, cannot be considered a semantically invariant transformation, unless the information specified is motivated by the context.* |

One must be cautious, because some seemingly simple transformations require deeper manipulations in order to respect grammatical agreements or because there are obstacles. These transformations are impossible to do with simple regular expressions applied on the surface form. They require to work more deeply at the syntactic level.

## 3.8 Technique 5 - Paraphrases generation using syntax trees transformations

The generation of paraphrases by syntactic tree transformations was the starting point of this study. The technique is directly inspired by the work of Michel Gagnon of École Polytechnique de Montréal who is contributing to this research [Gagnon & Da Sylva, 2005], [Zouaq, Gagnon & Ozell, 2010]. In these research, Dr Gagnon's team showed how to summarize texts by transforming dependency trees coming from a wide coverage morpho-syntactic parser. This rule-based parser was at the heart of the "Correcteur 101", a successful commercial French grammar checker [Bourdon et al, 1998], [Doll, Drouin, & Coulombe, 2005].

Parsing a sentence with a dependency grammar (DG) gives usually a tree whose nodes are the words of the sentence and the edges (links) the syntactic dependencies between the words.

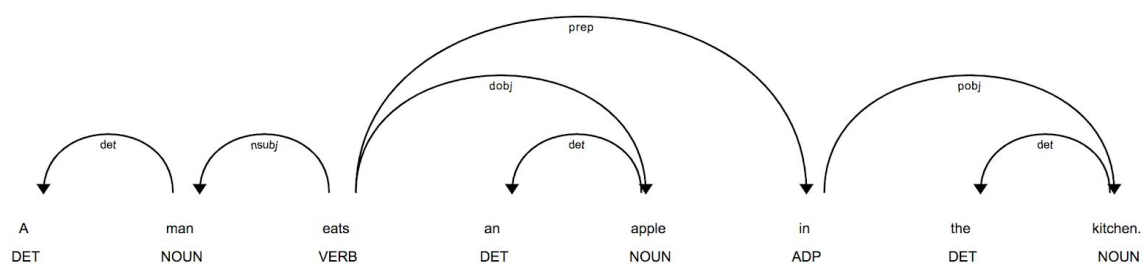

| Example of dependency tree produced by SyntaxNet |
| --- |
| A man eats an apple in the kitchen. |

```
root eats/eat/verb/proper_unknown/present/third/gender_unknown/singular
    nsubj man/man/noun/proper_unknown/tense_unknown/person_unknown/gender_unknown/singular
        det A/A/det/proper_unknown/tense_unknown/person_unknown/gender_unknown/number_unknown
    dobj apple/apple/noun/proper_unknown/tense_unknown/person_unknown/gender_unknown/singular
        det an/an/det/proper_unknown/tense_unknown/person_unknown/gender_unknown/number_unknown
    prep in/in/adp/proper_unknown/tense_unknown/person_unknown/gender_unknown/number_unknown
        pobj kitchen/kitchen/noun/proper_unknown/tense_unknown/person_unknown/gender_unknown/singular
            det the/the/det/proper_unknown/tense_unknown/person_unknown/gender_unknown/number_unknown
    p ./.punct/proper_unknown/tense_unknown/person_unknown/gender_unknown/number_unknown
```

Diagram drawn with the help of spaCy [Honnibal & Montani, 2017]

In a dependency tree, we can find the dependency between a verb and the noun which is at the head of its subject phrase, or the dependency between a noun and an adjective which modifies it. Let us take the opportunity to recall the pioneering work of the linguist Igor Mel'čuk from Université de Montréal on the dependency grammar (DG) [Bourdon et al, 1998], [Mel'cuk, 1988] and the Meaning-Text Theory [Wikipédia, Igor Mel'čuk].

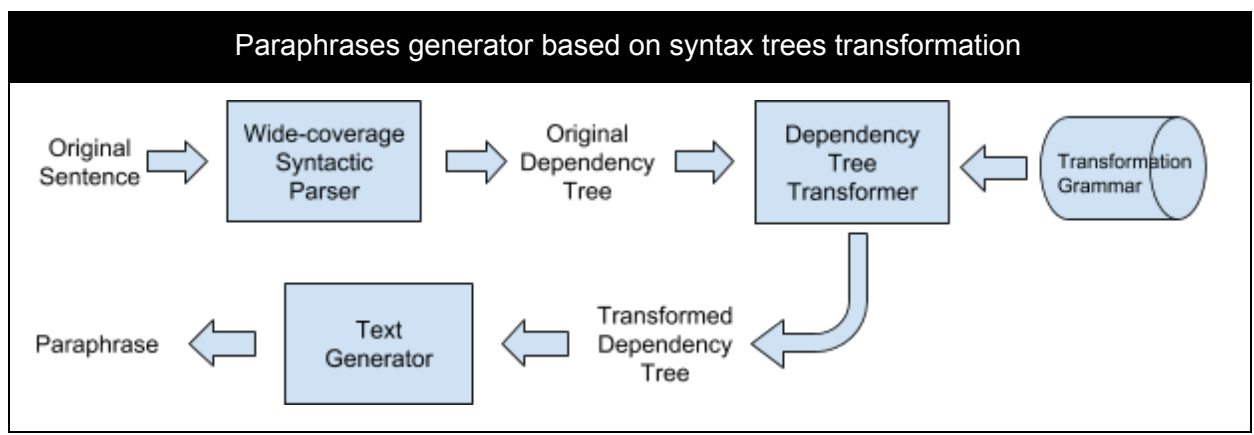

From the original sentence as input, a syntactic parser builds a dependency tree. Then the paraphrases generator transforms this dependency tree to create a transformed dependency tree guided by a transformation grammar. The transformed dependency tree is then used to generate a new surface form, ie paraphrase.

To work, the paraphrases generator needs a wide-coverage morphosyntactic parser that produces sentence analyzes in the formalism of dependency trees. Nowadays, there are powerful analyzers for more than fifty (50) languages that are available either in open source or in the form of online services.

The paraphrases generator proceeds sentence by sentence. Each sentence is parsed by SyntaxNet, an open source library from Google that uses deep learning techniques and the TensorFlow library. To be much effective, robust and scalable, the code is executed on the Google Cloud infrastructure using the Cloud Natural Language API [Google, 2018a]. The paraphrase generator's rule-based code was originally written in Prolog and then translated to Python in a tool called PolyPhrases.

| Comparison between PolyPhrases prototype and commercial tools | | | |
|---|---|---|---|
| A man eats an apple in the kitchen. To be or not to be. I took a train yesterday. | | | |
| A man eats an apple in the kitchen. to be or not to be. I caught a train last night. | A man eats an apple in the kitchen. Regarding life, what to think about it. I took a prepare yesterday. | A man fare Associate in Nursing apple within the room. To be or to not be. I took a train yesterday. | An apple is eaten by a man in the kitchen. To be or not to be. I took it yesterday. |
| commercial tool 1 | commercial tool 2 | commercial tool 3 | PolyPhrases prototype |

For the prototype, the rules of transformations, rather general, are built manually, without any assistance, guided by the typology of the paraphrases established by [Vila, Martí & Rodríguez, 2014]. The plan is to focus on the 20% of the rules that cover 80% of cases, acccording to the Pareto's engineering principle 20/80. Marie Bourdon, computational linguist at Coginov Inc Montréal, gave a very valuable help on that issue [Bourdon et al, 1998]. For the identification of additional rules and templates, it is possible to imagine the use of bootstrapping extraction techniques such as those described by [Androutsopoulos & Malakasiotis, 2010].

| Examples of semantically invariant syntactic trees transformations |
|---|
| *The transition from the passive verb form to the active verb form and vice versa is a semantically invariant transformation.* |
| *The replacement of a noun or a nominal group by a pronoun is a semantically invariant transformation.* |
| *The withdrawal of an adjective, an adverb, an adjectival group or an adverbial group is a semantically invariant transformation.* |

| Diagram of the transformation from active to the passive voice |
|---|

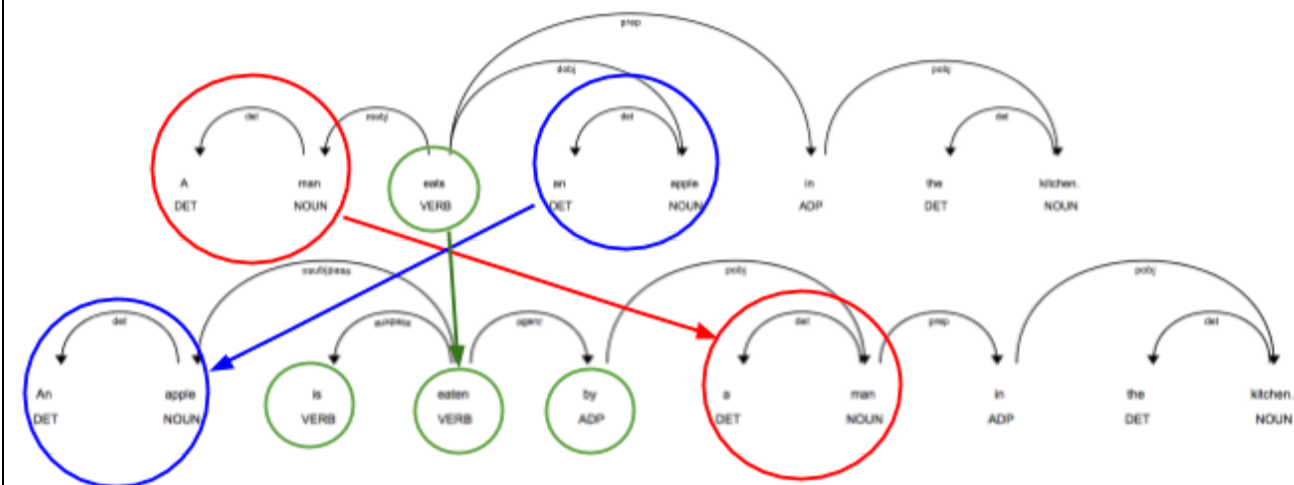

Diagram drawn with the help of spaCy [Honnibal & Montani, 2017]

Transformation from the active voice to the the passive voice of the phrase "A man eats an apple in the kitchen." The head of the dependency structure is the verb "eat". The transformation rule starts by exchanging the "man" subject group (in red) and the "apple" object group (in blue). Then the "eat" verb is modified (in green) to give a new dependency structure which once flattened generates the phrase "An apple is eaten by a man in the kitchen. "

## 3.9 Technique 6 - Paraphrases generation using back-translation

The study of the back-translation transformation was suggested by Dr Antoine Saucier, physicist, during a "discussion in a hallway" at the École Polytechnique de Montréal in spring 2015.

Back-translation is an old trick used to test the quality of a machine translation program. The back-translation consists in translating to the original language a text already translated from this language.

Historically, the first mention of the use of the retrotraduction under the term "round trip machine translation" to introduce variants into the text data can be found in an article by a team from King's College London presented at ISCOL 2015 [Lau , Clark & Lappin, 2015]. Back-translation is now an emerging technique employed by [Prabhumoye et al., 2018], a team bringing together people from Google and Facebook [Edunov et al., 2018] and [Wieting, Mallinson & Gimpel, 2017].

Thanks to the spectacular progress of neural machine translation (NMT) based on deep learning, the use of back-translation has become an interesting technique to experiment with. Recall that the recent advances in neural machine translation (NMT) largely originated from the pioneering work of the MILA lab and the Yoshua Bengio's team at Université de Montréal, which have then been perfected and industrialized by Google. Among MILA's contributions: neural models of language and the dense word vector idea behind Word2Vec [Bengio et al, 2003], the encoder / decoder architectures [Cho et al, 2014a], [Cho et al, 2014b] and the attention mechanism [Bahdanau, Cho & Bengio, 2014].

There is no such thing as a correct and unique translation of a sentence into a given language. In fact, there are still a large number of correct translations because of the immense combinational productivity of the natural language and the complexity of the real world. By definition, all these equivalent translations are paraphrases.

| Text data augmentation using back-translation |
| --- |
| *Good quality back-translation is a semantically invariant transformation.* |
| *Poor quality back-translation is not a semantically invariant transformation.* |

Like all other transformations, the back-translation transformation proceeds sentence by sentence. Each original phrase in English is translated a first time in a target language calling a translation module (locally or at distance on the cloud). A second translation is requested on each translation produced but this time to English, hence the concept of back-translation. These "back-translated" sentences are potential paraphrases that are filtered through different mechanisms to be retained or eliminated.

The result of the back-translation is filtered to recover the paraphrases. If the back-translation is identical to the original sentence, it is immediately rejected. If not, a similarity measure is done between the text of the retrotraduction and the original text. To be quite rigorous, it is rather to identify if the back-translation can be considered as a paraphrase of the original text. It is based on the observation that paraphrase is often quite similar to the original text [Dolan et al, 2004]. But this is not always the case and some good paraphrases will be rejected. At the end, the back-translation is preserved when its similarity with the original text is greater than a certain threshold which is empirically fixed.

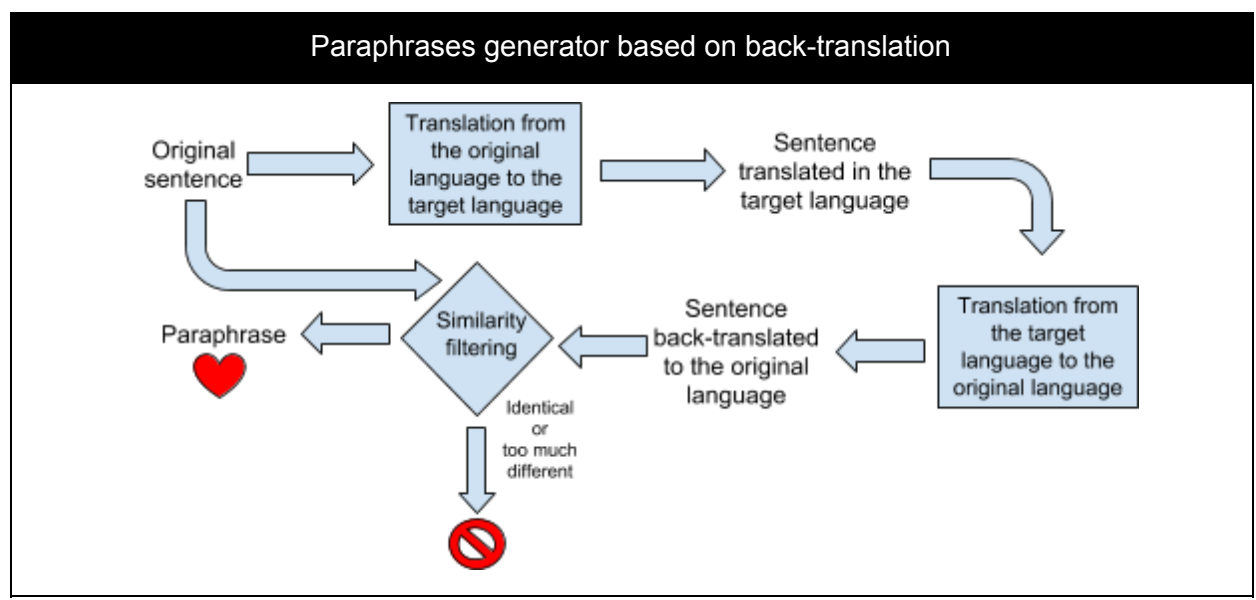

There exists a large number of methods to measure the similarity between two texts [Bär, Zesch & Gurevych, 2015], from the simple count of the number of words in common, through the cosine in a vector space, to the most recent methods based on neural networks that are able to identify paraphrases [Lan & Xu, 2018]. For reasons of rapid implementation and computational performance, we opted for a simple measure of the difference in length between the back-translated text and the original text. A similar technique is used by [Wieting, Mallinson & Gimpel, 2017].

A measure of gross similarity between an original text and its back-translation consists in simply counting their respective number of words. The threshold was quickly fixed by trial and error at 25%. All that could be fine-tuned doing more experiments.

Unlike [Edunov & al, 2018] or [Wieting, Mallinson & Gimpel, 2017], who use their own neural machine translation models, which require considerable computing resources, it has been more practical to use the online translation services from Google via Google Translate API. Recall that Google Translate, the online translation service of Google, uses neural machine translation (NMT) based on deep learning. Once registered and the developer's API key obtained, all the required Python code fit in one short cell of an iPython Notrebook.

## 4. EXPERIMENT

### 4.1 Tools

For our experiments, we mainly used **Python** tools available as open source software. **Scikit-Learn** for classical machine learning [Pedregosa et al, 2011]. **Pandas** for pre-processing and data manipulation [McKinney, 2010]. **Matplotlib** for graphing. **Keras** for deep learning experiment, a high level Python library for deep learning built on top of **TensorFlow** [Chollet, 2015], [Chollet, 2017 ]. **NLTK** library (Natural Language ToolKit) [Bird, Klein & Loper, 2009] which has an high level API for **Wordnet**. We start also using the **spaCy** library [Honnibal & Montani, 2017]. We used Google's **SyntaxNet**, a large-coverage analyzer based on dependency grammar [Petrov, 2016], [Kong et al, 2017]. Available in open source code and written with TensorFlow also in open source, SyntaxNet uses the Computational Natural Language Learning (CoNLL) formalism based on Universal Dependencies (UD) [Batista, 2017].

For practical reasons and good engineering practices (robustness, scalability, ease of deployment and cost), we opted to use services from the **Google Cloud Platform** [Google, 2018b]. Specifically, we use the syntactic SyntaxNet derived parsing service of the **Google Cloud Natural Language API** [Google, 2018c] and the automatic translation service, based on Google Translate, **Google Cloud Translation API** [Google, 2018d].

### 4.2 Task selection - text polarity prediction

The experiments are set up with a simple problem that involves a standard dataset and common architectures of deep neural networks. Thus, it will be easier to isolate the effect of the augmentation of textual data.

The choosen task is the polarity prediction, positive or negative, of an opinion. This is a supervised learning task where the training data set is labeled. Each example has been previously annotated with a positive or negative label. The purpose of the model is to predict the polarity of new opinions [Pang, Lee & Vaithyanathan, 2002]. For this task and with this precise corpus, the performances range from 70% for conventional learning algorithms to more than 90% for fine tuned deep neural networks.

Recall, that the objective is not to show that the different techniques of text data augmentation give models that are superior to the best approaches, but more modestly to demonstrate their feasibility. The objective is only to show that text data augmentation has a positive effect on the resulting models.

By focusing on the feasibility, we avoid the optimal tuning of the model that can be long, tedious and expensive in terms of computing resources. The tested models may still overfit, but the data augmentation should reduced it substantially.

## 4.3 Data - IMDB movie reviews

The data come from IMDB films, a database that could be described as a standard because widely used by the scientific community. In addition, sentiment prediction is a common example for deep learning in NLP.

More specifically, we used the dataset "polarity dataset v2.0" which has 1000 positive reviews and 1000 negative reviews extracted from the IMDB database. The dataset was made available in June 2004 by [Pang & Lee 2004]. The file has been uploaded to the Web [Cornel, 2004].

## 4.4 Experimental design

The experiments involve training different neural deep network architectures on the original text data and then on the augmented data using upon different augmentation techniques.

The experiment is divided in two (2) phases: 1) the data augmentation pre-processing,  2) the models training.

The models are then examined to see if there is any improvement or deterioration in the their prediction performance (accuracy). Training error and F1 measures have been computed as well.

Given our scarce computational resources, the experiments were limited to proof of concept alone. That is, to show that the proposed text augmentation techniques give better results than no augmentation. So the baseline is the result on the original data (no augmentation). Therfore much work remains to be done to explore each augmentation method in detail.

## 4.5 Textual data augmentation process

"Paraphrase" and "paratext" are two special objects or data structures created in order to describe the different possible combinations of parts (words for "paraphrase" and sentences for "paratext"). These are associative table or Python dictionary for each text that includes an index for each parts (part_1, part_2, ..., part_n) with which is associated a list of replacement parts with complementary information like the frequency or heuristic weight. The same algorithm is used to generate paraphrases from words or paratexts from sentences. Given the huge underlying combinatorics, the generation algorithm is sampling data using random and memorize each combination to avoid too much repetition. To handle large amounts of data, impossible to put in computer memory, one could easily virtualize the algorithm by storing data structures on disk.

All techniques process the texts sentence by sentence. The goal is to create a paratext object that will generate the number of variants of the original text that is required by the amplification factor. It is important to note that for these experiments, which only aimed at demonstrating feasibility, there were limited to the generation of five (5) texts for each original texts which represents a tiny part of all the possible combinations.

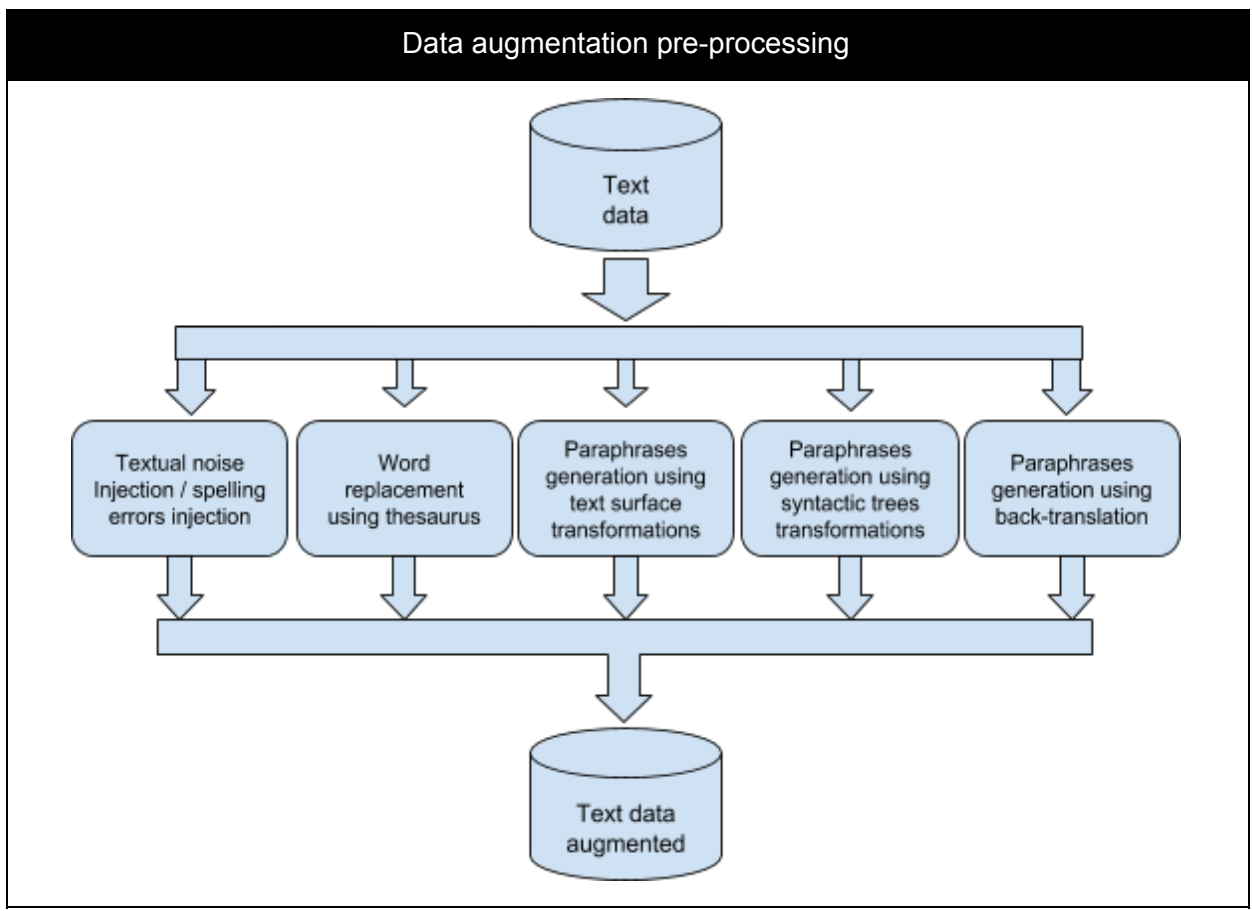

## 4.6 Models training process

The deep neural network architectures tested cover the three most commonly used types, multilayer perceptrons, MLPs, convolutional networks (CNN) and recurrent networks (RNN).

Specifically, we have experimented with a simple multilayer perceptron (MLP), a two (2) hidden layers MLP, a one (1) dimension convolutional neural network (CNN 1D), a long short-term memory recurrent network (LSTM}, and a bidirectional LSTM (biLSTM).

The neural nets were regularized only by adding more data and using early stopping (no external word embeddings, no L1, L2, no dropout, no hyperparameters tuning).

All the models are trained with a sufficient number of iterations (n_epochs = 10) so that the algorithm stops after a few iterations (generally 3 to 7) by early stopping. In fact, the algorithm very rarely goes to the limit of 10 iterations. In principle, continuing the iterations would not give anything and even worse, it could degrade the results.

As our models tend to overfit, we performed multiple evaluations and cross-validation. We have also implemented various mechanisms such as early stopping and the learning rate reduction available in the Keras tool [Coulombe, 2018].

To account for the variability of the results, we calculated an average over several executions (between 15 and 30, depending on the training duration) and also carried out a cross-validation with 5 and sometimes 10 folds.

## 5. RESULTS

Since accuracy is the standard criterion used for supervised classification tasks, the experiments measure the percentage of properly rated film critics or accuracy.

The published results below (table-1) are coming from the average over several executions (between 15 and 30, depending on the training duration). When there is a combination of many augmentation techniques, that corresponds simply to the addition of the datasets, not the application of a new augmentation technique to the data resulting from previous data augmentation technique [ie. 2+3 not 3(2()), 1+2+...+6, not 6(5(4(3(2(1())))))][2].

---

[2] Results for such successive applications of data augmentation are not available yet.

The baseline is the performance measured on the non amplified original data (nothing). This threshold varies according to the neural network architectures tested. For the single perceptron, the performance threshold to be exceeded was measured around 75%.

| **Observed Text Polarity Accuracy On Test Data**<br>(mean and standard deviation for 15 to 30 runs) | | | | | |
|---|---|---|---|---|---|
| | **Classical algorithm** | **Tested Neural Network Architecture** | | | |
| **Textual Augmentation Techniques** | XGBoost classical algorithm[3] | Simple Multilayer Perceptron | Multilayer Perceptron 2 hidden layers | LSTM recurrent NN | biLSTM recurrent NN |
| 0- Nothing **baseline** | 81.00% | 74.70% +/-2.36 | 76.40% +/-3.58 | 57.62% +/-2.68 | 54.25% +/-2.80 |
| 1- Textual Noise | **83.50%** **+/-%** | 83.10% +/-1.65 | 85.15% +/-0.69 | 65.17% +/-1.62 | 67.43% +/-1.30 |
| 2- Spelling Errors | 82.50% | 84.02% +/-2.56 | 80.70% +/-2.51 | 70.17% +/-1.87 | 67.50% +/-2.05 |
| 3- Synonyms Replacement | 81.50% | 80.73% +/-1.61 | 81.32% +/-1.02 | 69.97% +/-2.37 | 60.47% +/-2.29 |
| 4- Paraphrases Gen. RegEx | 81.50% | **85.40%** **+/-1.37** | 83.73% +/-1.36 | 67.77% +/-1.11 | 68.47% +/-1.60 |
| 5- Paraphrases Gen. Syntax Tree | 82.50% | 85.28% +/-0.77 | 82.85% +/-1.35 | 65.27% +/-2.38 | 69.17% +/-3.20 |
| 6- Back-Translation | 81.00% | 82.42% +/-0.97 | 82.20% +/-3.11 | 62.07% +/-2.45 | 67.10% +/-2.63 |
| 2 + 3 | 81.00% | 82.75% +/-2.29 | 80.82% +/-2.90 | 69.53% +/-2.94 | 68.00% +/-1.89 |
| 2 + 4 | 82.50% | 79.63% +/-1.89 | 80.95% +/-4.12 | 70.13% +/-3.06 | 66.47% +/-5.31 |
| 1+2+3+4+5+6 | 80.50% | 80.12% +/-3.77 | **85.92%** **+/-2.06** | **78.10%** **+/-2.28** | **75.87%** **+/-2.24** |
| Table-1 | | | | | |

**In all cases studied text augmentation increases the accuracy in a range 4.3% to 21.6%.**

---

[3] run only once

**Accuray vs Text Augmentation Techniques for different Neural Network Architectures**

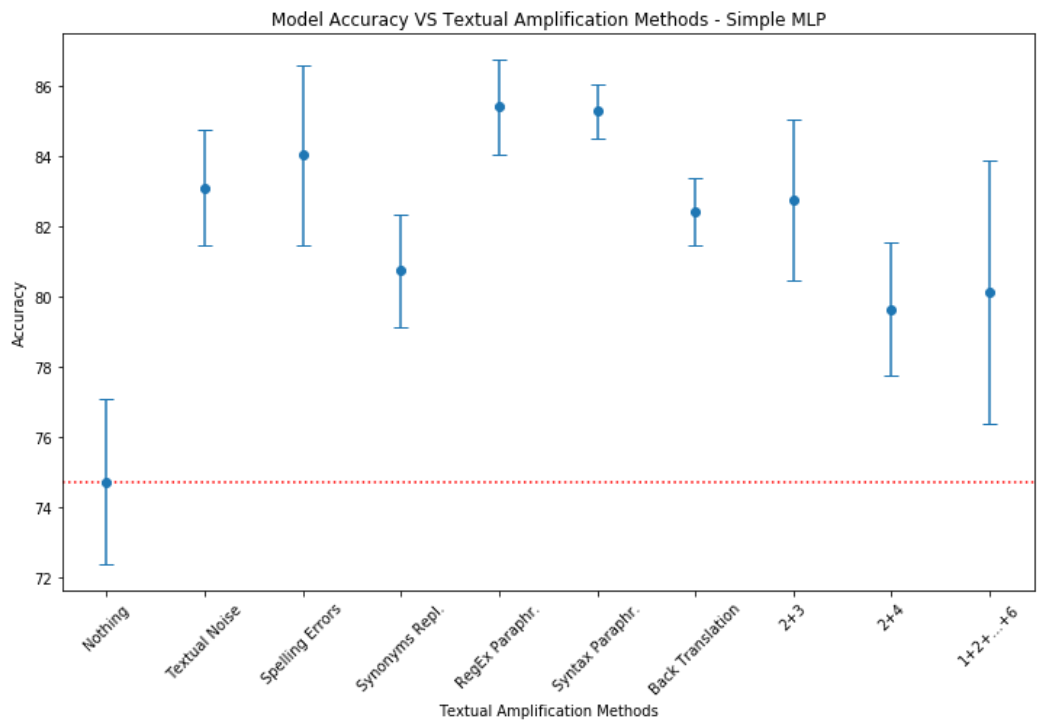

Simple MLP

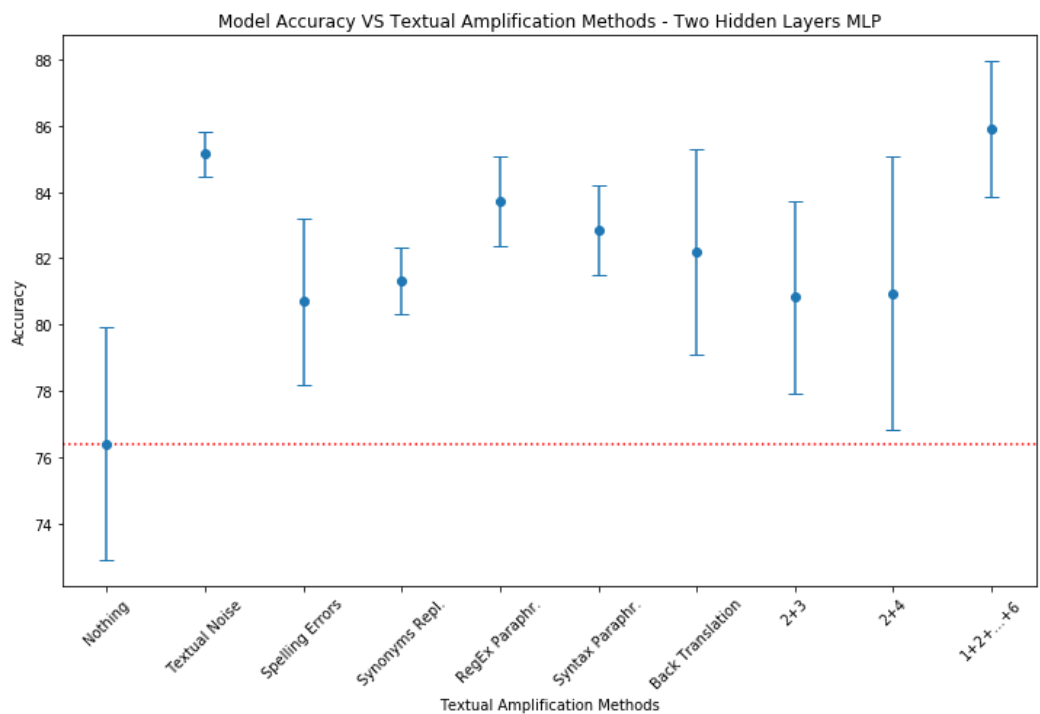

Two Hidden Layers MLP

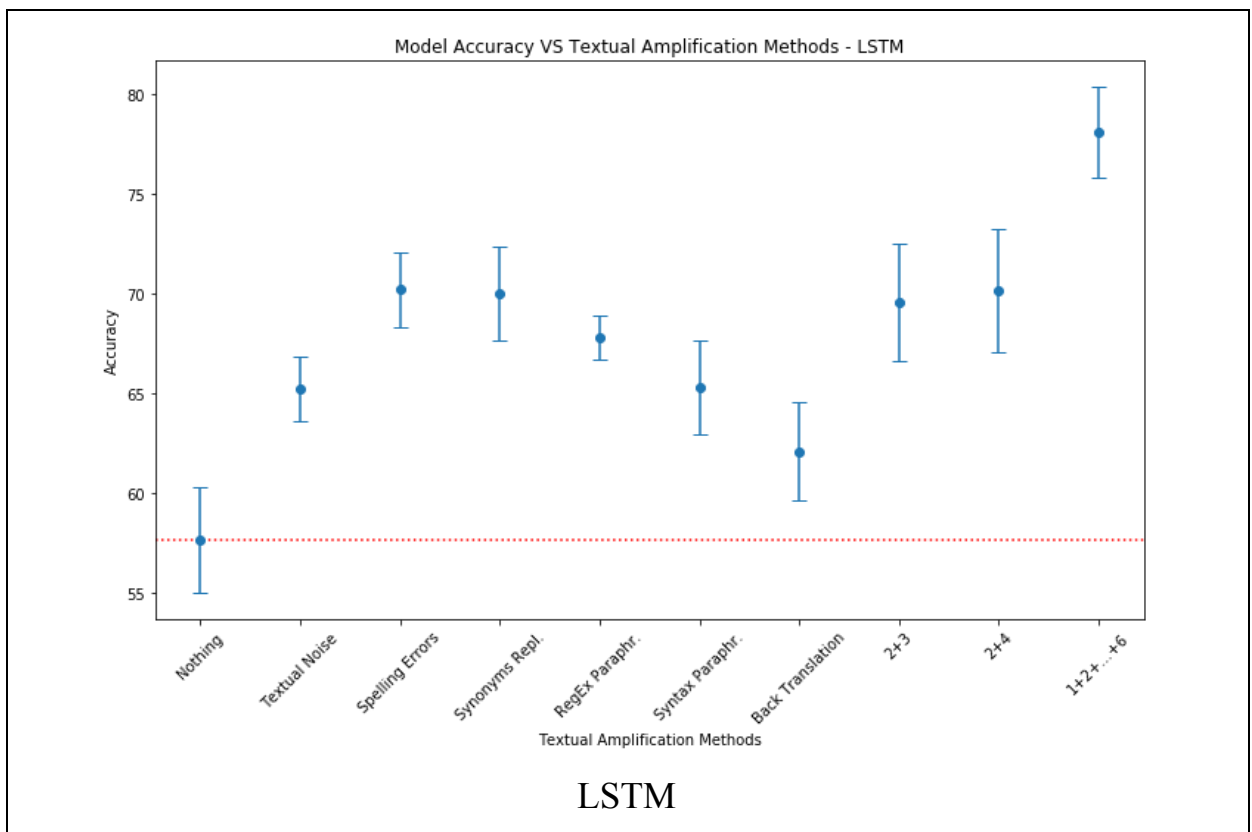

LSTM

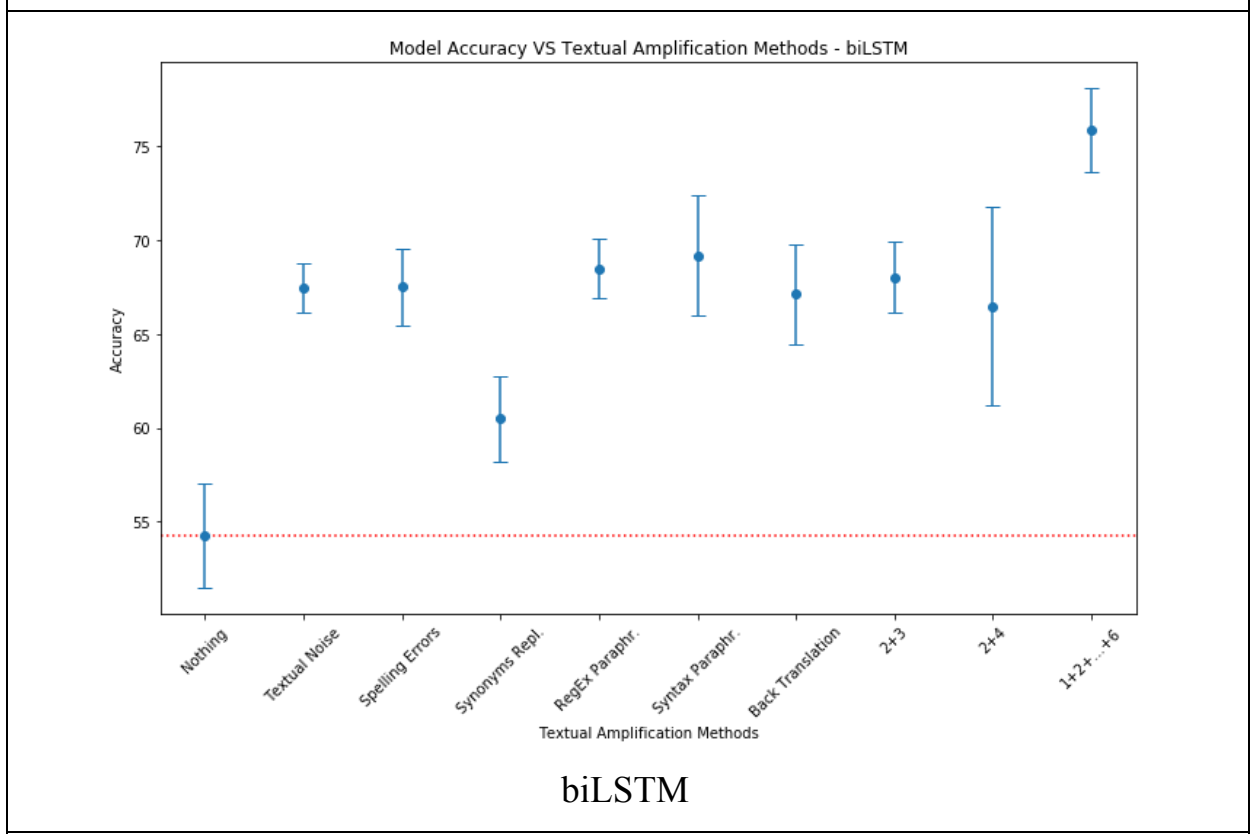

biLSTM

The results in the graphs above are presented with error bars which represent the standard deviation. This way of presenting information has the advantage of highlighting the effect of different textual augmentation techniques on the accuracy.

It is remarkable that for a recurrent network of LSTM type, one goes from models practically incapable of predicting anything (just a little better than the 50% due to chance) to models. able to predict in more than ⅔ cases and all this by injecting only textual noise.

Not surprisingly, recurrent neural networks, especially bi-directional biLSTM network, give the best results. It is especially interesting to note that they manage to make good use of the addition of the augmented data resulting from the different techniques.

We also tested with the classic XGBoost algorithm which is very effective in most situations that do not require deep learning. XGBoost algorithm is now considered the reference choice for small datasets. It is interesting to note that an excellent classical learning algorithm like XGBoost already manages to get the most out of the original data with 81%. XGBoost beats many of the neural networks but shows only a slight improvement of 2.5% with the augmented data. XGBoost struggles to improve with additional data because it has already managed to exploit most of the regularities / signals useful in the data and it has not enough capacity to learn new stuff from the augmented data.

In the defense of neural networks, there was no fine tuning. Neural networks were regularized only by adding data and using early stopping. So, no enrichment by external vector-words, no regularization L1, L2, no extinction of neurons (dropout), nor adjustment of hyperparameters.

## 6. DISCUSSION

### 6.1 Randomness of the results

An important observation with deep neural networks is the randomness of the results they produce. The same network trained with the same data can produce quite different results from one training to another. This randomness is necessary for their good functioning because it allows them among other things, to try different possibilities during training and to avoid getting stuck in a local minimum [Brownlee, 2018a]. It is important to maintain this variability when designing and developing a neural network.

Most of this variability is explained by the fact that the weights of the neural network are randomly initialized each time the training begins [Brownlee, 2018a]. In addition, it is well known that training a neural network using Stochastic Gradient Descent (SGD) rarely leads to the same result every time. To this, one must be added the random behavior of certain forms of regulation such as the extinction of neurons (dropout).

The most commonly means to reduce the effect of a particularly good or bad random seed is to train several models each with a different random seed and to combine their predictions according to the paradigm of ensemble learning methods [Brownlee, 2018a]. Similarly, cross-validation is a way to estimate the true performance of a model by averaging performance across multiple subsets of data.

This randomness explains why deep learning systems in production base their results on sets of models to give more consistent predictions.

## 6.2 Advantages

### 6.2.1 A software engineering dream comes true

The greatest benefits of the text augmentation techniques just briefly explored are from a practical and software engineering point of view.

Leveraging NLP cloud online services from well established web providers gives a lot of concrete and immediate advantages: availability, robustness, reliability, scalabilty. Furthermore, there are cheap, ready yo use and mostly available in a large number of languages.

They are also easy to implement and easy to use. Few lines of code are enough to call an online service and retrieve the results.

### 6.2.2 Comparison with emerging approaches

The main advantage of the proposed techniques over many new trends is that one does not try to generate meaningful sentences "ex nihilo" but rather by modifying existing sentences by means of semantically invariant transformations.

## 6.3 Drawbacks

The main disadvantage of some of the techniques explored is the amount of computation required which requires the use of cloud infrastructures.

There is also the dependence on translation and parsing services from private providers. For practical reasons, we opted to use services from the Google Cloud Platform. We outline alternative solutions to mitigate this dependence.

The preferred solution is to shop for services from other suppliers. For example, Amazon offers the Comprehend natural language processing solution [Amazon, 2018] and Microsoft its Microsoft Cognitive Services that include a translation service [Microsoft, 2018]. The different suppliers offer more or less free use quotas and more or less advantageous rates. It can be expected that in the medium term, all these online services will become commodities.

If you have an in-house IT infrastructure or a rented one, a second solution is to train your own neural translation model and your own morphosyntactic parser. Here, the main obstacle is to gather enough corpus of good quality.

Although SyntaxNet is rather difficult to install on a server or local machine, Google's open source license allows you to do that. Three ways are possible: docker image installation provided by Google [Docker, 2016], installation via a pre-configured server image (like AMI: Amazon Machine image) or installation on rented servers on a commercial cloud infrastructure from the source code of Google [Google, 2017], [Poddutur, 2018].

To replace Google Translate, it is still possible to train your own neural translation model from open source codes as [Edunov & al, 2018] and [Wieting, Mallinson & Gimpel, 2017] have already done.

## 6.3 The place of data augmentation in the data pipeline

There are specific situations where data augmentation is the only remedy available. These are fairly common situations where, despite the potential use of large pre-trained generic models for transfer learning, there is not enough data to specialize the model for a specific task.

## 6.4 Limits

The main limitation and criticism of this work is that the experiment was carried out only on a single task, moreover a very simple one which consists to predict the polarity of a text.

It is also important to note that the experiments were only intended to show the feasibility of different text augmentation techniques, mainly by lack of free access to a large computing infrastructure equipped with GPUs. As a result, the experiments were limited to an amplification factor of five (5), that is, five paratexts per original text, which represents a tiny fraction of all possible combinations, but enough to show the positive effect of data augmentation.

## 7. Conclusion

In conclusion, in the case of complex problems, the fact of not having enough data is a major obstacle in the use of deep learning. This is the "Big Data Wall".

This empirical work, conducted with limited computational resources, has shown that the use of different simple, practical and easy-to-implement text data augmentation techniques is likely to help cross the "Big Data Wall".

These techniques include textual noise injection, spelling errors injection, word replacement using thesaurus, and paraphrases generation using regular expression, paraphrases generation using syntactic tree transformations, and back-translation. The latter two are based on Cloud NLP APIs which are robust, scalable and easy-to-use.

The repeated 15 to 30-fold (and cross-validation) experiments with all these text data augmentation techniques have increased the accuracy of the results in a range of 4.3 to 21.6% on a simple standard task (supervised binary classification).

This work is more an engineering endeavour than a fundamental research. Nevertheless, the impact is likely to be significant for all practitioners, engineers and researchers seeking concrete and practical solutions to overcome the "Big Data Wall" of deep learning in NLP.

Source code[4]

---

[4] Code will be made available here: https://github.com/ClaudeCoulombe/TextDataAmplification/